\documentclass[11pt]{article}

\usepackage{hyperref}
\usepackage{url}
\usepackage{booktabs}
\usepackage{amsfonts}
\usepackage{nicefrac}
\usepackage{microtype}

\usepackage[margin=1in]{geometry}

\usepackage{amsmath,amssymb,mathtools,bm,bbm}
\usepackage[british]{babel}
\usepackage{graphicx}
\usepackage{algorithm,algorithmic}
\usepackage{pgfplots}
\pgfplotsset{compat=newest}
\usepgfplotslibrary{groupplots}
\usepgfplotslibrary{dateplot}

\usepackage[round]{natbib}

\title{Black-Box Inference for Non-Linear Latent Force Models}

\author{%
	Wil~O.~C.~Ward$^{1}$\thanks{Corresponding author [e: \texttt{w.ward@sheffield.ac.uk}]. $^1$Department of Computer Science, The University of Sheffield, UK; $^2$School of Computing, Newcastle University, UK; $^3$School of Mathematics, Statistics and Physics, Newcastle University, UK}%
	\and%
	Tom~Ryder$^{23}$%
	\and%
	Dennis~Prangle$^3$ \and%
	Mauricio~A.~\'Alvarez$^1$%
}
\date{{}}

\begin{document}
\maketitle 

\begin{abstract}
\noindent Latent force models are systems whereby there is a mechanistic model describing the dynamics of the system state, with some unknown forcing term that is approximated with a Gaussian process. If such dynamics are non-linear, it can be difficult to estimate the posterior state and forcing term jointly, particularly when there are system parameters that also need estimating. This paper uses black-box variational inference to jointly estimate the posterior, designing a multivariate extension to local inverse autoregressive flows as a flexible approximator of the system. We compare estimates on systems where the posterior is known, demonstrating the effectiveness of the approximation, and apply to problems with non-linear dynamics, multi-output systems and models with non-Gaussian likelihoods.
\end{abstract}

\section{INTRODUCTION}
Latent force models are a class of models that can be used to combine known mechanistic systems with non-parametric representations of unknown forces. Consider the example system, defined as a differential equation,
\begin{align}
	\alpha_0x(t) + \alpha_1\frac{\mathrm{d}}{\mathrm{d}t}x(t) +  \frac{\mathrm{d}^2}{\mathrm{d}t}x(t) = u(t),\label{eq:forced_spring}
\end{align}
which describes a forced mass-spring-damper system. In practice, the unknown forcing term, $u(t)$, may need to be estimated given only observations of the system state, $x(t)$. It may be that parameters $\alpha_0$ and $\alpha_1$ are also unknown, and need to be simultaneously estimated. Parametric forms for $u(t)$ may be used to estimate the state; however, placing a Gaussian process prior over $u(t)$ gives rise to the latent force model \citep{Alvarez2013}. The resulting model can be transformed into a regression problem, and solved a probabilistic solution to $x(t)$ and $u(t)$ can be inferred. An alternaive inference scheme using Kalman filtering and Rauch-Tung-Streibel smoothing was introduced by \citet{Hartikainen2010}, which infers the state sequentially. Latent force models have been shown as effective ways to approximate latent forcing terms, constraining the solution with the defined mechanistic dynamics of the system.

Solutions can only be inferred exactly in the case that the underlying differential equaton is linear. \citet{Alvarez2013} use linearisation of terms in non-linear models, which greatly simplifies the mechanistic aspect of the model, while \citet{Hartikainen2012} use non-linear filtering approaches for approximate inference. In the latter case, the results are effective for known parameters, but the sequential nature of the inference scheme leads to challenges in joint parameter estimation; such challenges are the inspiration for the work by \citet{Durrande2019}. Problems where the dynamics are partially known, but where one or more terms are non-linear, for example if $\alpha_1$ in \eqref{eq:forced_spring} were dependent on $x$ or $u$, occur in many areas, inclduing biomechanics \citep{Barenco2006}; population modelling \citep{Liu2010}, structural health monitoring \citep{Worden2018} and control systems \citep{Conte1999}. In the case where parameters are also unknown, sequential methods can struggle to perform joint inference effectively.

In this work, we seek to build a method that can jointly infer a system with particually known dynamics and some unknown forcing term; and model parameters, given noisy observations of only part of the model. We build a flexible approach using Bayesian variational inference to optimise a surrogate estimation of our posterior state. As in \citet{Ryder2018}, we use black-box variational inference and construct a neural network-based representation of the joint solution. To give our solution flexibility, we construct the variational distribution using normalising flows; designing a new architecture designed to model multivariate systems where high proportions of the state is unobserved.

The main contributions of this paper include building a simulation-based variational estimation for non-linear latent force models, using a sigma point method to propagate uncertainty in the loss term. We integrate Bayesian parameter estimation into the inference method, which includes marginal approximations of Gaussian process hyperparameters. The multivariate extension of the normalising flow is a novel contribution, designed to encode dependencies between latent dimensions based on the Markov properties of the underlying model. The approach is applied to a Gaussian process regression to demonstrate the ability of the normalising flow to approximate samples and apply model criticism, comparing the estimate with the known posterior to quantify the effectiveness of the approximation. We then apply the approach to simulated and real non-linear forced models to demonstrate the proposed approach for jointly estimating state, forcing term and parameters, and show how this can be extended to multivariate outputs. Further, we demonstrate that the method can be easily extended to problems with non-Gaussian likelihoods.

\section{BACKGROUND}
This section briefly reviews the related background of this work: discretising moments for Gaussian filtering of non-linear stochastic differential equations; and normalising flows in the context of this work.

\subsection{Continuous-Time Filtering}
Given a (non-linear) stochastic differential equation, describing some state variable, $x(t)$ driven by dynamics with coefficients $\alpha_i(x,t)$ and white noise, $w(t)$,
\begin{equation}
  \alpha_0(x,t)x(t) + \sum^n_{i=1}\alpha_i(x,t)\frac{\mathrm{d}^i}{\mathrm{d}t^i}x(t) = w(t),\label{eq:example_sde}
\end{equation}
we desire to regress the state on some observations, often noisly observed, e.g. $y\,|\,x \sim \mathcal{N}(x,\sigma^2)$. One such method is to use Bayesian filtering to calculate the posterior $p(\bm{x}\,|\,\mathbf{y})$ by forward propagating state-estimates and updating the approximation with observations.
\paragraph{Kalman-Bucy Filter} The Kalman filter is a widely recognised approach to solving state-space models conditioned on a set of observations \citep{Sarkka2013}. The Kalman filter relies on a discrete-time update, so the SDE must be discretised. If the system is linear and time-invariant, this can be performed exactly and the underlying state-space model is defined $\bm{x}_{k+1} = \exp((t_{k+1}-t_k)\mathbf{D})\bm{x}_{k} + \mathbf{Q}\bm{\varepsilon}_k$, where $\mathbf{D}$ is the companion matrix representing the $n$-order system as a first-order SDE, $\mathbf{Q}$ is derived from the steady-state covariance of the system, and $\bm{\varepsilon}_k \sim \mathcal{N}(\bm{0},\mathbf{I})$. Inference can be performed using standard Kalman filtering techniques \citep{Sarkka2013}.

\paragraph{State-Space Gaussian Processes}
The state-space model interpretation of a Gaussian process (GP) with finitely-differentiable covariance treats the regression as a continuous-time SDE that can be described in the form of \eqref{eq:example_sde}. The coefficients of the SDE, $\alpha_i$, are scalar and derived from the covariance function of a GP prior.

Given that the SDE representation of the Gaussian process is linear and time-invariant, due to the constant coefficients, it can be solved in the forward-direction exactly using the Kalman-Bucy filter. To calculate the full posterior given the data, a backwards pass using, e.g. the Rauch-Tung-Streibel smoother can be performed \citep{Hartikainen2010}.

The state-space representation of Gaussian processes also allows latent force models to be expressed as a joint companion system, combining the underlying system dynamics with the dynamics of the Gaussian process placed over the prior \citep{Hartikainen2012, Sarkka2018}.


\paragraph{Unscented Kalman-Bucy Filter}
Where the state to be inferred is subject to \emph{non-linear} dynamics, such that $\alpha_i$ in \eqref{eq:example_sde} are dependent on $x$, the posterior cannot be calculated exactly. There are, however, several approaches to state-space modelling of non-linear systems, including the extended Kalman filter, and sequential Monte Carlo filters \citep{Sarkka2013}. The unscented Kalman filter (UKF) is another such approach, using a set of so-called \emph{sigma points} to characterise the moments of the state estimate. Sigma points are individually propagated through the dynamics and can be combined by means of weighted sums to obtain approximations of the mean and covariance \citep{Julier1997}.

A continuous-time extension of the UKF was introduced in \citet{Sarkka2007}, defining continuous-time dynamics for predicting the moments, $\bm{m}(t)$ and $\bm{P}(t)$. For a given system, with dynamics $\bm{D}(\bm{x},t)$, diffusion term, $\bm{\Sigma}(t)$, and sigma points $\bm{\chi}$, the moments such that $\bm{x}(t) \sim \mathcal{N}(\bm{m}(t),\bm{P}(t))$ are defined
\begin{align}\begin{split}
  \frac{\mathrm{d}}{\mathrm{d}t}\bm{m}(t) &= \bm{D}(\bm{\chi},t)\bm{w}^{(m)}\\
  \frac{\mathrm{d}}{\mathrm{d}t}\bm{P}(t) &= \bm{\chi}(t)\mathbf{W}\bm{D}(\bm{\chi},t)^\top +  \bm{D}(\bm{\chi},t)\mathbf{W}\bm{\chi}(t)^\top + \bm{\Sigma}(t).
\end{split}\label{eq:unscented_moments}\end{align}
Weights $\bm{w}^{(m)}$ and $\mathbf{W}$ derive from the unscented transform and are defined in the construction of $\bm{\chi}$. The moments can be forward-solved using an ODE solver, such as a Runge-Kutta scheme \citep{Sarkka2007}. At observation times, the update-step of the discrete-time UKF can be used to update the estimate of $\bm{x}$, which can be used as an initial value for further prediction.

\subsection{Autoregressive Flows}
Normalising flows can be used to represent a probability distribution $q(\bm{x})$ as a differential transformation of some base density, e.g. $\bm{g} : \bm{z} \mapsto \bm{x}$, $\bm{z}\sim\mathcal{N}(\bm{0},\bm{I})$ \citep{Rezende2014}. The mapping, $\bm{g}$ must be invertible and differentiable, such that
\begin{equation*}
q(\bm{x}) = p(\bm{g}^{-1}(\bm{x}))\big|\bm{G}_{-1}(\bm{x})\big|,
\end{equation*}
where $\bm{G}_{-1}$ is the Jacobian of $\bm{g}^{-1}$.

The tractability of $|\bm{G}_{-1}|$ is a restriction on the choice of mapping; enforcing an autoregressive expression, $x_i = \bm{g}(z_1,\ldots,z_i)$ results in a Jacobian that is triangular and therefore the determinant is the product of its diagonal elements. This form of $\bm{g}$ gives rise to the concept of autoregressive flows \citep{Kingma2016}.

\paragraph{Inverse Autoregressive Flows}
\citet{Kingma2016} describe the mapping by reparametrising its inverse, transforming variables sampled from the base density by functions that represent the dependencies between dimensions. An inverse autoregressive flow (IAF) can be defined in terms of two transformations, $\bm{\mu}$ and $\bm{\sigma}$:
\begin{equation}
  x_i = \bm{\sigma}(z_1,\ldots,z_i)\cdot z_i + \bm{\mu}(z_1,\ldots,z_i)
\end{equation}

The log probability is therefore
\begin{equation}
  \log q(\bm{x}) = \log p(\bm{z}) - \sum\log\sigma_i,\label{eq:flow_density}
\end{equation}
where $\sigma_i = \bm{\sigma}(z_1,\ldots,z_i)$.

The shift and scale functions,  $\bm{\mu}$ and $\bm{\sigma}$ are often defined as outputs of a neural network encoding the autoregressive requirements of the system. For example, in \citet{Kingma2016} the authors make use of convolutional layers, such as ResNet \citep{He2016}.
%

\paragraph{Local IAFs}
An issue with IAFs as described is that they can become computationally expensive to compute for high dimensional $\bm{x}$ due to the large number of inputs to the shift and scale functions. \citet{Ryder2018b} introduce a local version of the IAF for approximating state-space models, in which the shift and scale depend only on the immediately preceding $r$ entries of the vector, termed the receptive field: $x_i = f(z_{i-r},\ldots,z_i)$. This approach is similar to that used in PixelCNN and WaveNet \citep{VanDenOord2016,Wavenet}, using causal convolutions with restricted kernel width.

\section{VARIATIONAL INFERENCE FOR NON-LINEAR LATENT FORCE MODELS}
We consider the approximate inference of the joint state posterior, $p(\bm{x}(t),u(t),\bm{\theta}\,|\,\mathbf{y})$ of a non-linear latent force model of the form
\begin{equation}
  \alpha_0(\bm{x},u,t;\bm{\theta})\bm{x}(t) + \sum^n_{i=1}\alpha_i(\bm{x},u,t;\bm{\theta})\frac{\mathrm{d}^i}{\mathrm{d}t^i}\bm{x}(t) = u(t),\label{eq:nl_lfm}
\end{equation}
where $u(t) \sim \mathcal{GP}(0, k(t,t'))$ is the prior placed over unknown force, and $\mathbf{y}$ are observations at times $\tau_j$, $j=1,\ldots,t$, such that $\bm{y}_j \sim \pi(\bm{h}(\bm{x}(\tau_j);\bm{\theta}))$, some likelihood conditional on an emission model $\bm{h}(\bm{x})$.

Given its intractability, we use variational Bayes to approximate the conditional posterior. We define a joint state vector $\bm{f}(t) = [\bm{x}(t), \mathrm{d}\bm{x}/\mathrm{d}t, \ldots,u(t_k), \mathrm{d}u/\mathrm{d}t,\ldots]^\top$ and construct a first-order companion SDE for \eqref{eq:nl_lfm} such that it can be written
\begin{equation}
  \frac{\mathrm{d}}{\mathrm{d}t}\bm{f}(t) = \bm{D}(\bm{f},t;\bm{\theta}) + \mathbf{L}\varsigma w(t),
\end{equation}
where $\bm{D}$ is the companion form dynamic, $\mathbf{L}$ is a column vector of zeros in all but the final row, which equals $1$; and $w(t)$ is a unit white noise process, and $\varsigma^2$ is the variance as derived from the state-space form of the GP prior.

For some finite-time mesh, $t_0,\ldots,t_T$, and observation times $\tau_1\,\ldots,\tau_N$, which for simplicity are contained within the mesh, the joint posterior
\begin{align}\begin{split}
  p(\bm{x}_{0:T},u_{0:T},\bm{\theta}\,|\,\bm{y}) \propto&\\
  p(\bm{\theta})p(\bm{f}_0\,|\,\bm{\theta})\prod^{T-1}_{k=0}&p(\bm{f}_{k+1}\,|\,\bm{f}_k,\bm{\theta})\prod^N_{j=1}p(\bm{y}_j\,|\,\bm{f}(\tau_j),\bm{\theta}),
\end{split}\label{eq:posterior}\end{align}
where subscripts denote discrete-time evaulations, e.g. $\bm{f}_k \triangleq \bm{f}(t_k)$.

\subsection{Variational Bounds}
Given the intractibility of \eqref{eq:posterior}, we construct a variational approximation of the posterior to jointly estimate the system state and parameters, in the form of $q(\bm{f},\bm{\theta}) = q(\bm{\theta})q(\bm{f}\,|\,\bm{\theta})$. To obtain the optimal approximation, we seek to find $q^\ast\in \mathcal{Q}$ to minimise the KL-divergence between our estimate and the posterior:
\begin{equation}
  \mathcal{KL}\!\left[q^\ast\,\|\,p\right] = \mathbb{E}_{\bm{f},\bm{\theta}\sim q}\!\left[\log q(\bm{f},\bm{\theta}) - \log p(\bm{x},u,\bm{\theta}\,|\,\mathbf{y})\right].
\end{equation}
Noting that $p(\bm{x},u,\bm{\theta}\,|\,\mathbf{y}) = p(\bm{x},u,\bm{\theta},\mathbf{y})/p(\mathbf{y})$ and that the evidence $p(\mathbf{y})$ does not depend on $q$, is equivalent to maximising
\begin{equation}
  \mathcal{L}(q) = \mathbb{E}_{\bm{f},\bm{\theta}\sim q}\!\left[p(\bm{x},u,\bm{\theta},\mathbf{y}) - \log q(\bm{f},\bm{\theta})\right]\label{eq:elbo}.
\end{equation}
To allow straightforward unbiased estimation of \eqref{eq:elbo}, we follow the example of \citet{Kingma2014}, \citet{Rezende2014}, and \citet{Titsias2014} by reparameterising $q$ and taking $\bm{f} = \bm{m}_f(\bm{\varepsilon}_f,\bm{\theta}; \bm{\phi}_f)$ and $\bm{\theta} = \bm{m}_\theta(\bm{\varepsilon}_\theta; \bm{\phi}_\theta)$. The functions $\bm{m}_f$ and $\bm{m}_\theta$ should be defined such that they are invertible mappings of random variables, $\bm{\varepsilon}_f, \bm{\varepsilon}_\theta \sim \mathcal{N}(\bm{0},\mathbf{I})$, and be parameterised by $\bm{\phi}_f$ and $\bm{\phi}_\theta$ respectively. Thus, the family of functions, $\bm{Q}$ representing $q$, are the functions $\bm{m}_f$ and $\bm{m}_\phi$, parameterised by variational parameters $\bm{\phi} = \{\bm{\phi}_f, \bm{\phi}_\theta\}$.

An unbiased estimate of $\mathcal{L}(q)$ can be obtained using $M$ Monte Carlo samples,
\begin{align}\begin{split}
  \mathcal{L}(q) \approx&\\
  \frac{1}{M}&\sum^M_{i=1}\log\left(p(\bm{\theta}^{(i)}p(\bm{f}^{(i)}\,|\,\bm{\theta}^{(i)})p(\mathbf{y}\,|\,\bm{f}^{(i)},\bm{\theta}^{(i)})\right)\\
  &\qquad-\log\left(q(\bm{\theta}^{(i)})q(\bm{f}^{(i)}\,|\,\bm{\theta}^{(i)})\right),
\end{split}\label{eq:unbiased_elbo}\end{align}
where $\bm{f}^{(i)} = \bm{m}_{f}(\bm{\varepsilon}^{(i)}_f,\bm{\theta}^{(i)};\bm{\phi}_f)$ and $\bm{\theta}^{(i)} = \bm{m}_\theta(\bm{\varepsilon}^{(i)}_\theta;\bm{\phi}_\theta)$ for independent samples $\bm{\varepsilon}^{(i)}_f$, $\bm{\varepsilon}^{(i)}_\theta$ from $\mathcal{N}(\bm{0},\mathbf{I})$.

To maximise $\mathcal{L}(q)$ with respect to $\bm{\phi}$, we use a stochastic gradient optimisation algorithm, approximating $\nabla_\phi\mathcal{L}$ with an unbiased estimate of sample gradients, calculated using automatic differentiation of \eqref{eq:unbiased_elbo} \citep{Ranganath2014,Ryder2018}.

\subsection{Discretisation}
In calculating $\mathcal{L}$, we must approximate the marginal of our joint system state $p(\bm{f}\,|\,\bm{\theta}) = \prod p(\bm{f}_{k+1}\,|\,\bm{f}_k,\bm{\theta})$. To do so, we assume that transition is (approximately) normal, and build moments using the prediction dynamics of the continuous-time unscented Kalman filter, described by \eqref{eq:unscented_moments},
\begin{equation}
	\bm{f}_{k+1}\,|\,\bm{f}_k,\bm{\theta} \sim \mathcal{N}(\bm{m}_{k+1}, \bm{P}_{k+1}).
\end{equation}
We construct the discrete-time predictions using Euler's method, with step size $\Delta_t = t_{k+1}-t_k$ \citep{Griffiths2010}. The update steps for the mean and covariance of the transition density are defined
\begin{align}
	\bm{m}_{k+1} &= \bm{f}_k + \Delta_t\bm{D}(\bm{\chi}_k,t;\bm{\theta})\bm{\omega}^{(m)}\\
	\bm{P}_{k+1} &= \bm{\Sigma} + \Delta_t\bm{U}(\bm{\chi}_k,t;\bm{\theta}),
\end{align}
where
\begin{equation}
	\bm{U}(\bm{\chi},t;\bm{\theta}) = \bm{\chi}\mathbf{W}\bm{D}(\bm{\chi},t;\bm{\theta})^\top +\bm{D}(\bm{\chi},t;\bm{\theta})\mathbf{W}\bm{\chi}^\top + \bm{\Sigma},
\end{equation}
and $\bm{\Sigma} = \tilde{\bm{L}}\bm{Q}_c\tilde{\bm{L}}^\top$. $\bm{Q}_c$ represents the $m\times m$ steady-state covariance of the state-space GP prior of $u$, and $\tilde{\bm{L}}$ is a $d \times m$ masking term to map $\bm{Q}_c$ to $d\times d$, where $d$ is the dimension of the joint state.

Unscented transform sigma points $\bm{\chi}_k$ and corresponding weight terms $\bm{\omega}^{(m)}$ and $\bm{W}$ are defined in matrix-form and built from base density $\mathcal{N}(\bm{f}_k, \bm{\Sigma})$ \citep{Sarkka2007}. Sigma points are defined
\begin{align}
	\small\bm{\chi}_k = \begin{bmatrix}\bm{f}_k & \ldots & \bm{f}_k\end{bmatrix} + \begin{bmatrix}\bm{0} & \sqrt{(d+\eta)\bm{\Sigma}} & - \sqrt{(d+\eta)\bm{\Sigma}}\end{bmatrix},
\end{align}
with weight terms defined
\begin{align}\begin{split}
	\omega^{(m)}_0 &= \eta(d+\eta)^{-1}\\
	\omega^{(c)}_0 &= \eta(d+\eta+1-\alpha_\chi^2+\beta_\chi)^{-1}\\
	\omega^{(m)}_i &= \omega^{(c)}_i = (2d+2\eta)^{-1}
\end{split}\\
\begin{split}
\bm{\omega}_{-1} &= \mathbf{I} - \begin{bmatrix}\bm{\omega}^{(m)}&\ldots&\bm{\omega}^{(m)}\end{bmatrix}\\
\mathbf{W} &= \bm{\omega}_{-1}\text{diag}[\bm{\omega}^{(c)}]\bm{\omega}_{-1}^\top,
\end{split}
\end{align}
where $\eta = \alpha_\chi^2(d+\kappa_\chi) - \kappa_\chi$ is the unscented scaling term and $\alpha_\chi$, $\beta_\chi$, and $\kappa_\chi$ are hyperparameters of the transform \citep{Julier1997}.

\subsection{Multivariate Masking of Local IAF}
As in both \citet{Kingma2016} and \citet{Ryder2018b}, we build $q(\bm{f}\,|\,\bm{\theta})$ as a hierarchy of inverse autoregressive flows to create a flexible approximation density. Because we are dealing with multivariate states, we design a novel variant for vector-valued temporal states.

First, the system state is flattened such that each entry corresponds to a single dimension, and the hierarchical flow transformation is constructed such that each layer acts only on a single dimension in the state. Successive layers alternate updates for each dimension, as motivated in \citet{Dinh2016}. A receptive field mask is used to enforce locally temporal dependencies, and each flow layer takes in the output of the previous layer, the model parameters, $\bm{\theta}$, and a local feature vector, $\mathcal{D}$.

\begin{figure}[t]
	\centering
	\def\svgwidth{\columnwidth}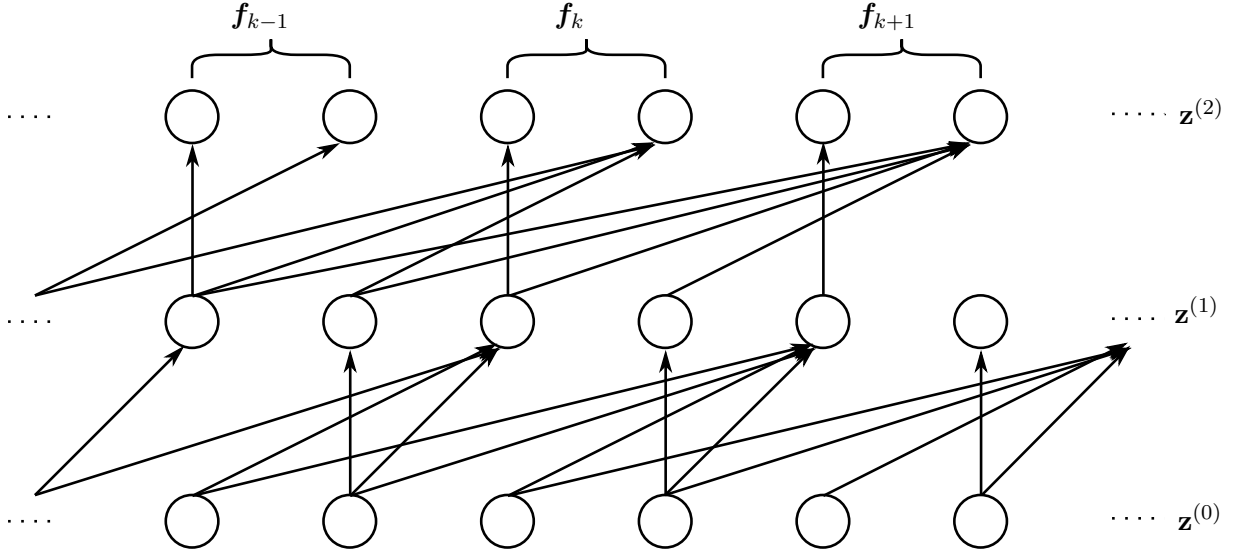
	\caption{Example architecture of multivariate masking of local inverse autoregressive flows, with state dimension $d=2$ and receptive field width $r=2$}
	\label{fig:dependency_graph}
\end{figure}

A demonstration of the flow dependencies is shown in \figurename{}~\ref{fig:dependency_graph}. Formally, the flow hierarchy can be defined for each element of the flattened joint state vector rolled out over time as such:
\begin{align}\begin{split}
	z^{(0)}_i &= \varepsilon_i\\
	z^{(l)}_i &= (\delta^l_i
    \sigma_i + (1-\delta^l_i))z^{(l-1)}_i +  \delta^l_i\mu_i,
\end{split}\label{eq:mv_local_iaf}\end{align}
where $\bm{\varepsilon} \sim \mathcal{N}(\bm{0},\mathbf{I})$, $\delta^l_i = 1 \text{ if } (i \text{ mod }l = 0) \text{ else } 0$ and
\begin{equation}
[\mu_i,\sigma_i] = \text{\textsc{AutoregressiveNN}}(\mathbf{r}^{(l)}_i\odot\bm{z}^{(l-1)},\bm{\theta},\mathcal{D}).
\end{equation}
The autoregressive neural network here encodes the conditional depencencies of the system. The receptive mask, $\mathbf{r}^{(l)}_i$ is a binary vector that masks the local receptive field. In practise, the receptive mask is enforced using causal 1-D convolutions with kernel width equal to $r\cdot d$, where $d$ is the state dimension and $r$ is the receptive field's width. The autoregressive neural network is constructed from multiple layers of convolutions with batch normalisation between layers. Details of the architecture are provided in Algorithm~\ref{alg:autoregressive_network}.

The final flow, $z^{(N)}$ is used to generate an estimate for $\bm{f}$, with $\bm{f} = \mathrm{flatten}^{-1}(\bm{h}(\bm{z}^{(N)}))$, where $\mathrm{flatten}^{-1} : \mathbb{R}^{dT\times 1} \to \mathbb{R}^{d\times T}$ is the inverse transformation to map the flattened vector back to its multivariate form. We also define $\bm{h}$ to be some optional bijector to enforce constraints on $\bm{f}$, for example $\bm{h}(\cdot) \triangleq \log(\exp(\cdot)+1)$, the softplus operator, can be used to enforce positivity of $\bm{f}$.

The log probability of the multivariate extension to local IAF is thus
\begin{align}\begin{split}
	\log q(\bm{f}) = \frac{T}{2}(\bm{\varepsilon}^\top\bm{\varepsilon} &+ \log 2\pi) +  \sum^T_{i=1}\sum^N_{l=1}\delta^l_i\sigma_i + (1-\delta^l_i) \\&+ \log|\bm{H}_{-1}(\bm{f})|,
\end{split}\end{align}
where $\bm{H}_{-1}$ is the Jacobian of the inverse of $\bm{h}$.

\paragraph{Local Features}
The features for the inverse autoregressive flow represent the additional input data to the model, including parameters and observation data. The flattening of the latent dimension is mirrored in the flattening of the observation data, with latent dimensions receiving $0$ as input.

For each element of a given flow, $\bm{z}^{(l)}_i$, the feature vector, $\mathcal{D}$ consists of: the current discrete-time point, $t_k$; the time until the next observation, $\tau_j - t_k$, such that $\tau_{j-1} < t_k \leq \tau_j$; the next observation, $\bm{y}_j$; and a binary mask indicating with value $1$ that there is an observation at $\tau_k$ \emph{and} that the current corresponding state dimension is not latent.

The feature vector is concatenated with the base sample of the flow before passing to the autoregressive network used to generate shift and scale terms, $\bm{\mu}^{(i)}$ and $\bm{\sigma}^{(i)}$. A sample from the parameter distribution is encoded with a densely-connected multilayer perceptron and the output is added to the first layer of the autoregressive network.

\begin{algorithm}[b!]
	\caption{$l^\text{th}$ \textsc{AutoregressiveNN}$(\bm{z}^{(l-1)},\bm{\theta},\mathcal{D})$}
	\label{alg:autoregressive_network}
	\begin{algorithmic}
		\STATE $\xi^{(0a)} \gets \mathrm{\textsc{conv1d}}(\bm{z}^{(l-1)}, \mathcal{D})$
		\STATE $\xi^{(0b)} \gets \mathrm{\textsc{dense}}(\bm{\theta})$
		\STATE $\xi^{(1)} \gets \mathrm{\textsc{elu}}(\xi^{(0a)} + \xi^{(0b)})$
		\FOR{$i = 2\ldots n_\ell$}
			\STATE $\xi^{(i)} \gets \mathrm{\textsc{batchnorm}}(\mathrm{\textsc{conv1d}}(\mathrm{\textsc{elu}}(\xi^{(i-1)})))$
		\ENDFOR
		\STATE $[\bm{\mu},\bm{s}] \gets \mathrm{\textsc{conv1d}}(\xi^{(n_\ell)})$
		\STATE $\bm{\sigma} \gets \mathrm{\textsc{softplus}}(\bm{s})$
		\STATE \textbf{return} $[\bm{\mu}, \bm{\sigma}]$
	\end{algorithmic}
\end{algorithm}

\subsection{Joint Parameter Estimation}
For the estimation of model parameters, we use a separate variational distribution $q(\bm{\theta})$, to be optimised as part of the maximisation of \eqref{eq:unbiased_elbo}. A simple family of distributions for $q$ is the family of mean-field Gaussian approximations. Here, we sample each parameter from an independent multivariate Gaussian, parameterised by $\bm{\phi}_\theta = \{\bm{\mu}_\theta, \bm{s}_\theta\}$. In this case, $\bm{m}_\theta(\bm{\varepsilon}^{(i)}_\theta;\bm{\phi}_\theta)$ is defined such that $\bm{\theta}^{(i)} = \mathrm{diag}[\bm{s}_\theta]\bm{\varepsilon}^{(i)}_\theta + \bm{\mu}_\theta$. Constraints on elements of $\bm{\theta}$ can also be applied, for example by placing the variational distribution of $\log \theta_i$ if it is constrained to be positive.

For the models presented in this paper, empical results demonstrated that the mean-field approach is sufficient for parameter estimation, particularly when there is known independence, such as between parameters of the input GP and of the system dynamics. However, such an approach can often be a poor choice for more complex dependencies \citep{Blei2017}, so alternative approaches may be used, e.g. masked autoregressive density estimation \citep{Germain2015}.

\section{RELATED WORK}

Physically-inspired inference of unknown systems with Gaussian processes can be considered in the process convolution interpretation of multi-output GPs \citep{Alvarez2011}, which consider the system as an integral problem with a shared latent GP describing the dependencies. This approach has been used for latent force models \citep{Alvarez2013}, and introduced to non-linear dynamics in \citet{Lawrence2007} and \citet{Titsias2009}, more recently being generalised to non-linear systems using a series approximation \citep{Alvarez2019}. The interpretation of latent force models as state-space models has been applied to non-linear problems in \citet{Hartikainen2010} using the unscented Kalman filter, but does not apply joint parameter estimation. Incorporating non-Gaussian likelihoods in the state-space approach to GP regression has been discussed in \citet{Nickisch2018}.

Approximation methods for ODEs and SDEs with fully unknown dynamics that use GPs to approximate some part of the system include gradient-matching GP regression fits, as in \citet{Wenk2018}, or approximating the phase and diffusion matrix of non-linear oscillators with GPs given only observations \citep{Heinonen2018, Yildiz2018}. Recent works investigating parameter estimation in stochastic systems with known dynamics includes approaches using variationa; inference \citep{Ryder2018, Binkowski2018}, and MCMC \citep{Abbati2019}.

This paper addresses the combined problem of partially known dynamics with unknown parameters using autoregressive neural networks. Similar simulation-based inference methods for sequential data include the sequential neural likelihood \citep{Papamakarios2019}, which presents a likelihood-free inference model with masked autoregressive flows \citep{Papamakarios2017}. A related approach for low-dimensional state-space models was introduced in \citet{Ryder2018b}.

\begin{figure}[t!]
\includegraphics[width=\columnwidth]{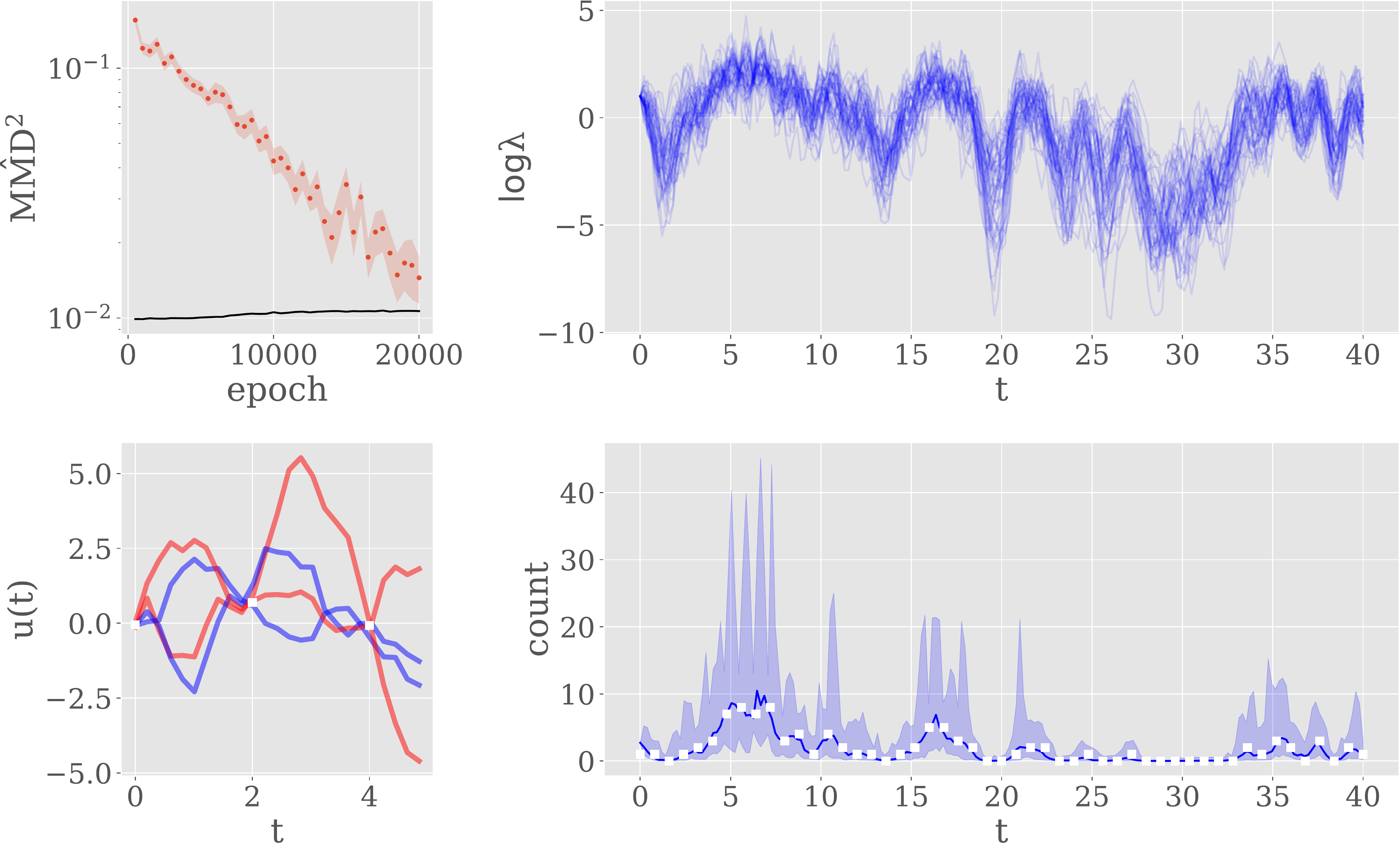}
\caption{Maximum mean discrepancy ($\textsc{mmd}^2$) scores comparing samples from the variational approximation of GP with Mat\'ern-\nicefrac{3}{2} covariance with samples the true posterior, plotted against training epoch [top left]. The black line indicates the threshold under which the null hypothesis can be accepted with 95\% confidence. Two samples from the true posterior (red) and variational approximation (blue) conditioned on some observations (white) [bottom left]. Samples form a variational approximation of a latent GP conditioned on count data [top right], with predictive density plotted with 95\% confidence intervals (shaded)}\label{fig:gp_plots}
\end{figure}
\section{MODEL CRITICISM}

A special case of the model defined in \eqref{eq:nl_lfm} is an SDE with the deterministic dynamics of a Mat\'ern GP, and the forcing term has a GP prior with white-noise covariance. The resulting system would be equivalent to a state-space GP. Thus, we can infer the posterior exactly for evaluation of the proposed approximate inference approach.

\paragraph{Mat\'ern covariances} The Mat\'ern family of covariances are finitely differentiable and, as such, can be represented exactly as SDEs \citep{Sarkka2019}. For a GP with half-integer Mat\'ern covariance, e.g. Mat\'ern-\nicefrac{3}{2}, the dynamics can be easily represented as a stochastic LFM:
\begin{equation}
	\lambda^n x(t) + \sum^{n}_{i=1}\begin{pmatrix}n\\i\end{pmatrix}\lambda^{n-i}\frac{\mathrm{d}^i}{\mathrm{d}t^i}x(t) = u(t),\label{eq:matern}
\end{equation}
where $u(t) \sim \mathcal{GP}(0, v(2\lambda)^{2n-1}(n!)^2/(2n-2)!\delta_{tt'})$, with parameters $v$, the variance; and $\lambda =\ell^{-1}\sqrt{2n-2}$, where $\ell$ is the so-called length-scale.

\paragraph{Maximum Mean Discrepancy}
We use a kernel-based two-sample test \citep{Gretton2012} to compare samples from the variational approximation with samples from the true posterior. Mapping the corresponding samples to a reproducing kernel Hilbert space using a Gaussian kernel, we apply the two-sample test using maximum mean discrepancy (\textsc{mmd}) as a similarity metric between the approximate and true posterior.

The $\mathrm{\textsc{mmd}}^2$ value for the approximation of Mat\'ern-\nicefrac{3}{2} GP fit is shown against the number of training epochs in \figurename{}~\ref{fig:gp_plots}. The figure shows that as training increases, the quality of approximation increases. The black solid line along the lower part of the plot indicates the threshold at which the null hypothesis, that the two distributions are the same, can be rejected with $95\%$ confidence. The trend observed thus indicates that with training the approximation is tending to being statistically indistinguishable in such a test. For visual reference, two samples from the true posterior and approximation, further demonstrating their similarity.

\section{EXPERIMENTAL RESULTS}
This section details experiments for problems with intractible posteriors, demonstrating use-cases and the flexibility of the proposed approach.

\paragraph{Non-Gaussian Likelihoods}
We apply the regression problem representing latent GP with Mat\'ern-\nicefrac{3}{2} covariance as in the previous section, but condition on a simulated set of count data to demonstrate that the approach can be easily extended to problems with non-Gaussian likelihoods. The regression problem is defined such that $\bm{f}(t)$ is the joint state that has dynamics defined in \eqref{eq:matern} with $n=2$
\begin{equation*}
	p(y_j\,|\,\bm{f}(\tau_j)) = \mathcal{P}ois(y_j\,|\,\exp(f_1(\tau_j)).
\end{equation*}
Samples from the variational distribution are shown in \figurename{}~\ref{fig:gp_plots}, along with the approximated predictive density which shows that the approximation can capture the main features and uncertainty of the system. The negative log predictive density (NLPD) for the fit using the proposed approach is $-0.12218$, versus $-0.16576$ for a GP fit with a Laplace approximation \citep{Rasmussen2006}.

\begin{figure}[t!]
	\centering
	\def\svgwidth{\columnwidth}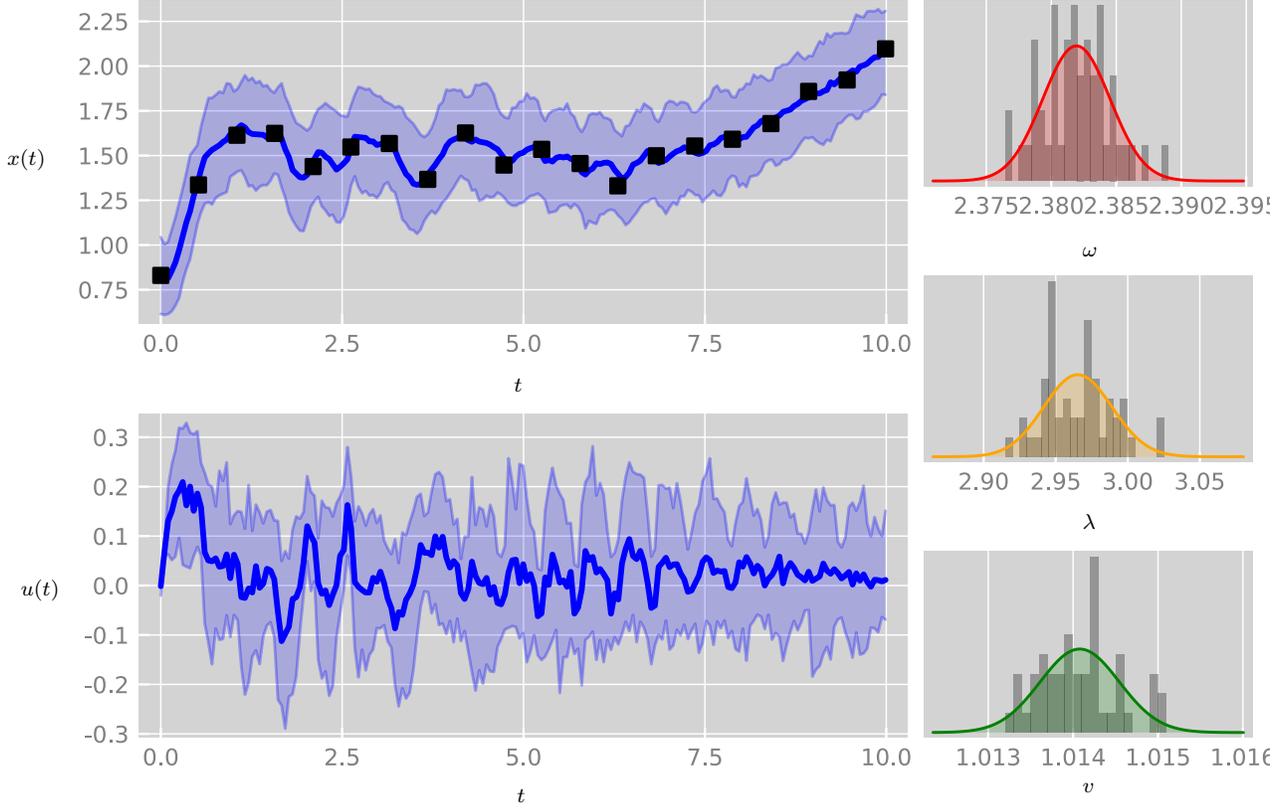
	\caption{Latent state, $x(t)$, and forcing term, $u(t)$, estimates using the proposed approach, conditioned on noisy observations of $x(t)$ indicated by black squares. The shaded area indicates the central 95\% sample quantile about the mean (solid blue). Plots on the right indicate the estimated marginals of the parameters and binned samples}
	\label{fig:forced_ode}
\end{figure}

\paragraph{Toy problem}
In this example, we demonstrate inference of a latent input function on a non-linear ODE using the proposed method. We consider a toy example with sinusoidal dynamics of an observable state, and place a Mat\'ern-\nicefrac{3}{2} GP prior over the unknown input:
\begin{equation}
	\frac{\mathrm{d}}{\mathrm{d}t}x(t) = -\frac{2}{3}\sin(\omega x(t)) + u(t)\label{eq:forced_ode}
\end{equation}
Parameters from both the latent state dynamics and input covariance, $\bm{\theta} = \{\omega, v, \lambda\}$ are jointly inferred with a mean-field variational distribution $q(\bm{\theta})$. The approximate posterior was conditioned on observations generated from a sample solution to \eqref{eq:forced_ode}, with additive Gaussian noise.

\figurename{}~\ref{fig:forced_ode} shows the inferred posterior state and latent forcing term inferred using the variational approach. The marginals for the inferred parameters are also shown. We observe that the undelying forcing term has largest effect in the early parts of the model, $t < 5.0$, which corresponds to deviation from the fixed dynamics observed in $\mathbf{y}$.

\begin{figure*}[t!]
	\centering
	\def\svgwidth{\textwidth}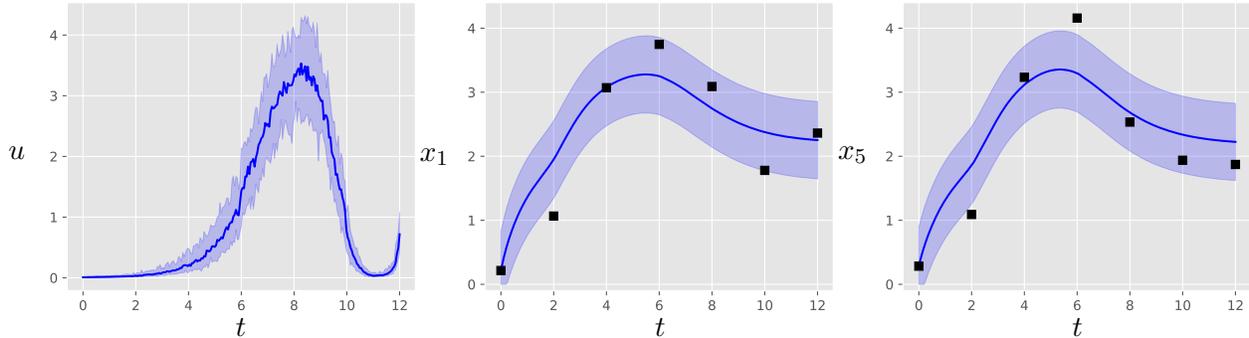
	\caption{Inferered transcription factor concentration (left) and predicted gene expression for \textsc{ddb}2 and p26 sesn1 (right). Black squares indicate measured gene expression, and shaded region represented the $95^\text{th}$ sample quantile about the mean }\label{fig:gene_expression}
\end{figure*}

\paragraph{Gene Expression}
In the final experiment, we consider a multi-output system using real-world data. We consider the transcriptional regulation model in \citet{Barenco2006}: an ODE describing the dynamics of target gene expression that are regulated by an unobserved transcription factor, $u(t)$. For each gene in the dataset, $x_d(t)$, the dynamics can be described as
\begin{equation}
	\frac{\mathrm{d}}{\mathrm{d}t}x_d(t) = a_d - b_dx_d(t) + s_d\frac{u(t)}{\gamma_d + u(t)}.
\end{equation}
We assume the gene expressions are observed with some additive noise, $\sigma^2_y = 0.25$ and place a GP prior over $\log u(t)$, as in \citet{Titsias2009}. Model parameters for each output include the basal transcription rate, $a_d$, decay rate, $b_d$, and sensitivity, $s_d$, which are all unknown. $\gamma_d$ is the Michaelis constant defined for each specific gene. Additional shared parameters are those of the GP covariance function, characterising $u(t)$.

We perform inference of the gene expression problem for $5$ observed genes: \textsc{ddb2}, \textsc{bik}, \textsc{tnfrsf10}b, \textsc{CI}p1/p21, and p26 sesn1. A Mat\'ern-\nicefrac{3}{2} GP prior is placed over the log transcription factor term, $\log u(t)$, and a mean-field approach is used to estimate parameters for each gene $\{a_d,b_d,s_d\,|\,d=1,\ldots,5\}$, and GP paremters, $\lambda$ and $v$. A softplus bijector was placed on the output paths representing $x_d$ to enforce positivity.

Plots showing the inferred (unscaled) transcription factor concentration, $u(t)$, and two of the gene expression states: \textsc{ddb2} and p26 sesn. We observe that the estimates of observable states, $x_d$, capture the observations and general form.

We observe that the proposed approach can easily extend to multidimensional states, and represent multi-output problems. In practice, this is similar to process convolution representation of such problems \citep{Alvarez2013}, however we are able to perform inference over non-linear models, while jointly estimating system parameters.

\section{CONCLUSIONS}
We present an approach for jointly inferring the parameters and state of non-linear ODEs with unknown forcing terms using a simulation-based autoregressive variational approximation. The approach can effectively simulate Gaussian process samples and infer both observed and latent states constrained to partially known dynamics systems. The method could be extended to non-linear SDEs with unknown input terms using the same proposed approach. We further demonstrate that the model can represent multi-output systems and models with non-Gaussian likelihoods.

There are some limitations to the approach, such as a tendancy to over-confidence; this can be observed in \figurename{}~\ref{fig:gene_expression}, where we observe narrow error bars in the latent GP for lower values of $t$. This might be fixing certain parameters of the GP, such as the variance, or deriving some term to penalise deviation from the prior more strictly.

We have proposed a new approach to approximating non-linear latent force models as a filtering problem, using a new multivariate masking architecture of local inverse autoregressive flows to handle dependencies between observable and latent state dimensions. The joint model can effectively learn parameters in such problems with batch gradient descent; a challenge for sequential approaches such as Kalman filtering or sequential Monte Carlo due to the need to roll-out gradients through time. Further, the proposed inference method is scalable to more complex variational distributions over the parameter space; and we can in principle extend the state-dimension arbitrarily, with any number of latent and observed terms.

\paragraph{Acknowledgements} WOCW and MAA have been financed by the Engineering and Physical Research Council (EPSRC) Research Project EP/N014162/1. MAA has also been financed by the EPSRC Research Project EP/R034303/1. TR is supprted by the EPSRC Center for Doctoral Trainig in Cloud Computing for Big Data (EP/L015358/1).

\bibliographystyle{abbrvnat}

\bibliography{draft}

\clearpage
\appendix
\section*{Companion Matrices for Differential Equations}
We briefly describe companion matrices for turning $n$-order linear SDEs into first-order by representing the system as a linear operator on an augmented state variable.

Consider the 2-order system
\begin{equation*}
  a_0x(t) + a_1\frac{\mathrm{d}}{\mathrm{d}t}x(t) + a_2\frac{\mathrm{d}^2}{\mathrm{d}t^2}x(t) = w(t)
\end{equation*}
Define a new variable, $z = \mathrm{d}x/\mathrm{d}t$, and substitute into the above equation:
\begin{align*}
  a_0x(t) + a_1z(t) + a_2\frac{\mathrm{d}}{\mathrm{d}t}z(t) = w(t)
\end{align*}
This is now the 1-order system:

\begin{align*}
  \frac{\mathrm{d}}{\mathrm{d}t}x(t) &= z(t)\\
  \frac{\mathrm{d}}{\mathrm{d}t}z(t) &= -\tilde{a}_0x(t) - \tilde{a}_1z(t) + a_2^{-1}w(t),
\end{align*}
where $\tilde{a}_0$ and $\tilde{a}_1$ are $a_0/a_2$ and $a_1/a_2$ respectively.

We can write this using a joint state, $\bm{x}(t) = \big[x(t)\  z(t)\big]^\top$:

\begin{align*}
  \frac{\mathrm{d}}{\mathrm{d}t}\bm{x}(t) = \begin{bmatrix}0 & 1\\-\tilde{a}_0 & -\tilde{a}_1\end{bmatrix}\bm{x}(t) + \begin{bmatrix}0\\a_2^{-1}\end{bmatrix}w(t)
\end{align*}
The companion matrix for linear $n$-order systems scales in a similar manner, with the final matrix consisting of a final row of scalars and off-diagonal band with value $1$. The extension to non-linear systems is a straightforward extension here, simply replacing the matrix with a vector-valued function.


\section*{Unscented Transform}
The unscented transform is a means for propagating a random variable, $x$ through a non-linear functional, $f$, by optimally sampling about the mean and propagating each sample through $f$ and combining the results as a weighted sum \citep{Julier1997}. These so-called sigma points are defined as:
\begin{align*}
	\chi_i &=\begin{cases} \mathbb{E}[x] & i = 0\\
	 \mathbb{E}[x] + \big[\sqrt{(n+\eta)\text{cov}[x]}\big]_i & i = 1,\ldots,n\\
	 \mathbb{E}[x] - \big[\sqrt{(n+\eta)\text{cov}[x]}\big]_{n-i} & i = n+1,\ldots,2n\end{cases}.
\end{align*}
Note that $[\cdot]_i$ indicates the $i^\text{th}$ column of a matrix. $n$ describes the dimension of the random variable $x$ and $\eta$ is a scaling parameter, defined such that $\eta=\alpha_\chi^2(n+\kappa_\chi)-n$.

The unscented transform consists of transforming each sigma point, $\gamma_i = f(\chi_i)$ and constructing a weighted sum. The approximation of $y = f(x)$ is given by $y \sim \mathcal{N}(\mu,\Sigma)$, where
\begin{align*}
	\mu &= \sum^{2n}_{i=0}\omega^{(m)}_i\gamma_i\\
	\Sigma &= \sum^{2n}_{i=0}\omega^{(c)}_i\big(\mu-\gamma_i\big)\big(\mu-\gamma_i\big)^\top.
\end{align*}
The weights are defined by
\begin{align*}
	\omega^{(m)}_0  &= \eta(n+\eta)^{-1}\\
	\omega^{(c)}_0 &= \eta(n+\eta+1-\alpha_\chi^2+\beta_\chi)^{-1}\\
	\omega^{(m)}_i &= \omega^{(c)}_i = (2n+2\eta)^{-1},
\end{align*}
where $\alpha$, $\beta$, and $\kappa$ are hyperparameters controlling the spread of sigma points. There are a number of reported settings for values of these hyperparameters, one such is $\alpha_\chi=1$, $\beta_\chi=0$, and $\kappa_\chi=n$ \citep{Julier1997}. These are the values used in this paper.

\section*{Implementation}
The implementation was written in TensorFlow, using TensorFlow probability. All experiments were optimised using Adam with a learning rate of 5e-3. For additional numerical stability, gradient clipping was performed based on the global norm.

The unscented filtering updates were implemented using the sigma-point dynamics as described in \citet{Sarkka2007}.

The number of flows for each experiment was set to $2d$, where $d$ is the latent state dimension.

\end{document}